\documentclass[letterpaper, 10 pt, conference]{ieeeconf}  % Comment this line out if you need a4paper

\IEEEoverridecommandlockouts                              % This command is only needed if 
\usepackage{amsmath} % assumes amsmath package installed
\usepackage{amssymb}  % assumes amsmath package installed
\usepackage{mathrsfs}
\usepackage{graphicx} %use graph format
\usepackage{epstopdf}      %Figure
\usepackage{cite}
\usepackage{subfigure}
\usepackage{float}
\usepackage{xcolor}                                                         % you want to use the \thanks command
\usepackage{algorithm}
\usepackage{algorithmic}

\title{\LARGE \bf
NEARL: Non-Explicit Action Reinforcement Learning for Robotic Control
}

\author{Nan Lin$^{1^*}$, Yuxuan Li$^{1^*}$, Yujun Zhu$^{1}$, Ruolin Wang$^{1}$,
  Xiayu Zhang$^{1}$, Jianmin Ji$^{1}$, \\Keke Tang$^{2}$, Xiaoping
  Chen$^{1}$,  and Xinming Zhang$^{1}$% <-this % stops a space
  \thanks{$^{1}$Nan Lin,  Yuxuan Li,  Yujun Zhu, Ruolin Wang,  Xiayu Zhang,  Xiaoping
    Chen, Xinming Zhang are with  University of Science and Technology of China, Hefei, 230026, China.
    {\tt\small xinming@ustc.edu.cn}}%
  \thanks{$^{2}$Keke Tang is with the Cyberspace Institute of Advanced
    Technology, Guangzhou University, Guangzhou 510006, China.}%
  \thanks{*These authors contributed equally to this work.}%
}

\begin{document}
%\begin{CJK*}{UTF8}{song}
\maketitle
\thispagestyle{empty}
\pagestyle{empty}

%%%%%%%%%%%%%%%%%%%%%%%%%%%%%%%%%%%%%%%%%%%%%%%%%%%%%%%%%%%%%%%%%%%%%%%%%%%%%%%%
\begin{abstract}
% Traditionally reinforcement learning methods predict the next action base on the current state.
% However in real world applications, action is low-level in terms of controlling.  
% Directly applying actions on control systems or robots may lead to unexpected behaviors. 
% In this paper, we propose a novel reinforcement learning framework without explicit action. Our meta policy directly manipulates the next optimal state and actual actions are produced by the inverse dynamics model. To stabilize the training process, adversarial learning and information bottleneck are integrated into our framework. Moreover, our method is able to directly utilize frequently encountered state-only demonstrations for imitation learning. Plus, prior domain knowledge can be added, hence improving the training efficiency. We tested our algorithm and the experiment results show that our framework and algorithm are reliable.

Traditionally, reinforcement learning methods predict the next action based on
the current state. However, in many situations, directly applying actions to
control systems or robots is dangerous and may lead to unexpected behaviors
because action is rather low-level. In this paper, we propose a novel
hierarchical reinforcement learning framework without explicit action.
Our meta policy tries to manipulate the next optimal state and actual action is
produced by the inverse dynamics model. To stabilize the training process, we integrate adversarial learning and information bottleneck into our framework. Under our framework, widely available state-only demonstrations can be exploited effectively for imitation learning. Also, prior knowledge and constraints can be applied to meta policy. We test our algorithm in simulation tasks and its combination with imitation learning. The experimental results show the reliability and robustness of our  algorithms.

\end{abstract}

%%%%%%%%%%%%%%%%%%%%%%%%%%%%%%%%%%%%%%%%%%%%%%%%%%%%%%%%%%%%%%%%%%%%%%%%%%%%%%%%
%%%% MAIN BODY %%%%
\section{INTRODUCTION}
Reinforcement learning has a broad application prospect in decision-making and control tasks.
In some sophisticated missions, RL-based methods have exhibited diverse behaviors and even outperformed human experts. \cite{silver2017mastering, akkaya2019solving}
Yet, reinforcement learning can be of low efficiency with expensive costs and we cannot ensure its reliability and safety. Hence its application in the real world is limited.
Besides, many real-word control system architectures are hierarchical. In such systems, only the state of the system can be manipulated, and the actions are output from a low-level controller.
It is hazardous to perform actions straightly through reinforcement learning, which may trigger unexpected behaviors.
In order to limit agents' behavior, recent work manages to apply designated constraints to reinforcement learning \cite{achiam2017constrained,miryoosefi2019reinforcement}.

\begin{figure}[!thb]
  \centering
  \includegraphics[width=8.2cm]{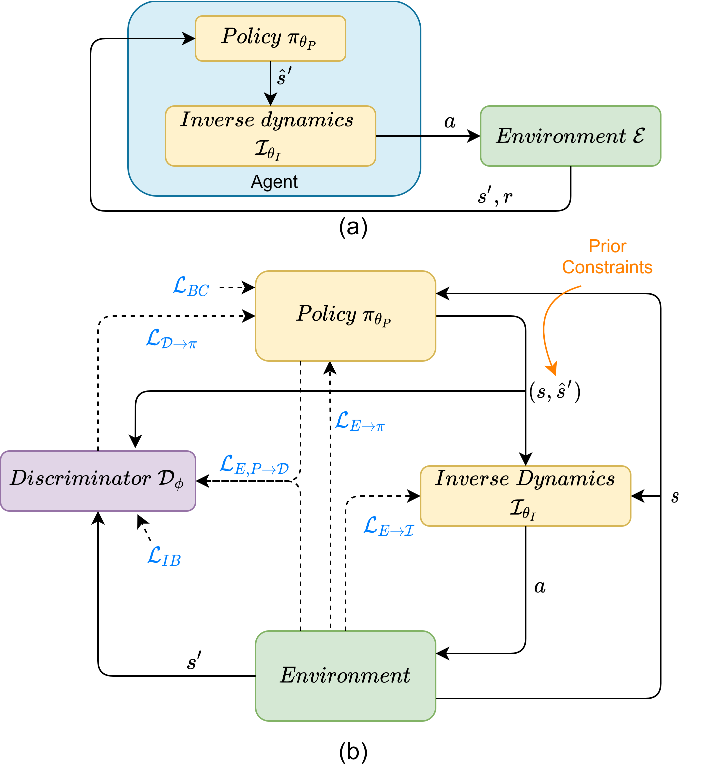}
  \caption{(a) Original NEARL framework. Meta policy maps the current state to the next one and inverse dynamics model outputs the actual action based on the predicted next state and the current state.
    (b) Flowchart of algorithm $\rm NEARL-PID^2$. Dashed line represents different loss functions and solid lines represent interaction between the environment, policy, inverse dynamics and discriminator. Optional prior constraints can be applied to the predicted states.}
  \label{fig:model}
\end{figure}

Imitation learning is able to obtain skills from observing demonstrations from humans or experts. \cite{argall2009survey, hussein2017imitation}
Combining reinforcement learning and imitation learning is proven to improve the learning efficiency
as well as ensuring the diversity of agents' behavior.
In common practices, actions of the demonstration are needed for the learner to imitate expert policy. However, in certain cases, expert's demonstrations with actions cannot be easily obtained. State-only demonstration is more of the case, like human kinetic trajectories obtained by motion capture system \cite{peng2018deepmimic} and online video data \cite{peng2018sfv}, which is not compatible with supervised learning methods.

In this paper, we propose a novel Non-Explicit Action Reinforcement Learning (NEARL) framework. 
Different from other reinforcement learning or imitation learning methods, state-to-state meta policy is learned directly, i.e., we predict the next optimal state based on the current state. Hierarchical architecture is adopted, and a low-level inverse dynamics model is used to infer the action interacting with the environment. A GAN-like framework is added to restrict policy to choose only reachable states. An information bottleneck is also deployed to filter redundant information in state space to stabilize the training process. 

A state-to-state policy has expectedly many advantages over its traditional
state-action pair counterpart. For instance, traditional controllers such as
trajectory planner or offline demonstration data may not contain any
action, while our approach can still utilize these data conveniently. Additionally, the state-to-state meta policy inherently integrates an
environmental model, which enables our method not only to be used in one step
Markov Decision Process (MDP) but also sequence learning to solve complex
problems like long horizon tasks.

In summary, the contribution of this paper is:
\begin{enumerate}
    \item  A Non-Explicit Action Reinforcement Learning framework that helps to combine other control methods and effectively utilize prior domain knowledge.
    \item  A novel control algorithm integrated with GAN-style method and information bottleneck to stabilize the training process.
    \item  Validation experiments that prove the feasibility and robustness of our framework and algorithms.
\end{enumerate}

The rest of this paper is organized as follows: In
Section II, we review the related work. In
section III, we present our NEARL framework and its background and notations.
In section IV, we proposed a set of control algorithms based on NEARL framework.
The Experiment section contains verification of our framework and effectiveness
of the algorithm is shown. Performance with
state-only demonstration and comparisons with other imitation learning methods are also demonstrated. In the end, conclusions and future work are summarized.

\section{Related Work}
\subsection{Model-based Reinforcement Learning}
Model-based reinforcement learning attempts to learn a dynamics model from data for later planning or policy search.
Several supervised learning methods are available to learn an environmental model like time-varying linear model
\cite{levine2013guided,levine2014learning}, random forests\cite{hester2013texplore}, nearest neighbors\cite{jong2007model} and non-parametric Bayesian methods \cite{deisenroth2011pilco}. For high-dimensional control, advanced approaches are adopted, such as deep auto-encoders \cite{wahlstrom2015pixels}, Bayesian neural networks \cite{depeweg2017learning} and variational inference \cite{moerland2017learning}.

Another crucial problem is how to utilize the model. A common practice
is to update policy through planning on the model. For example, Dyna\cite{sutton1991dyna}
learns policy from both real and simulated experience; more recently,
Guided Policy Search \cite{levine2013guided, levine2014learning} uses guiding
samples to optimize the policy search.

Apart from policy updating, the model can also play a part
in generating action interacting with the environment by combining with Model Predictive Control
theory, including iLQR\cite{watter2015embed}, direct optimal control
\cite{chua2018deep}, model predictive path integral control\cite{williams2017information}, etc.

Compared to traditional Model-free RL, Model-based RL has advantages in data efficiency, for it being expensive to sample in the real world due to time cost and limited life expectancy of robots. Model-based RL can utilize data effectively, but it heavily depends on the accuracy of the model \cite{sutton2018reinforcement}.
Unfortunately, it is not easy to guarantee the model accuracy in data-driven methods.

\subsection{Imitation learning from Demonstration}
Currently behavior cloning \cite{giusti2015machine,bojarski2016end} and inverse reinforcement learning \cite{bain1995framework,argall2009survey} are two widely-used imitation learning methods.
Behavior cloning is a
class of imitation algorithms where supervised learning is performed to learn policies that imitate the experts, by minimizing the action prediction error
in demonstrations. Behavior cloning comes with simplicity and efficiency, but sometimes it may suffer the distribution
shift problem \cite{codevilla2019exploring}. 
Apart from using the demonstration
data to  learn a map from states to actions directly, inverse reinforcement learning is motivated to learn a hidden reward
function to reduce the problem into traditional RL.
Some latest practices involve adversarial learning to
minimize the discrepancy between imitator and demonstrator.
\cite{ho2016generative,peng2018variational, torabi2018generative}.

Imitation learning improves the learning effciency, but these approaches arbitrarily require agents have access to
trajectories with expert actions, which are not always observable.
Such obstacles can be overcome with recently proposed imitation learning from demonstrations (IfD) or
imitation learning from
observations (IfO)\cite{torabi2019recent}.

In reward shaping approaches of IfD, the reward can be hand-crafted to encourage the agent to match the current state with reference motion clip at each step
\cite{peng2018deepmimic} or used to measure differences between policy and expert in embedding space \cite{sermanet2018time,aytar2018playing}.
Unfortunately, such a reward shaping method requires sufficient samples, similar to the traditional reinforcement learning method. 

Inverse dynamics model (IDM) exists as another option. It maps from
state-transitions $(S_{t}, S_{t+1})$ to actions. Therefore, the IDM is capacitated to complete state-only
demonstrations with inferred actions, and then the problem can be reduced to regular
imitation learning problems \cite{radosavovic2020state,
  torabi2018behavioral}. However, its training
depends on the current policy distribution with likely instability.

Meanwhile, generative adversarial learning can also be adopted in inverse dynamics model methods. Merel et al. propose an IfO algorithm to produce human-like movement patterns from limited
demonstrations consisting of the only partially observed state features \cite{merel2017learning}. Torabi et
al. develop a method named generative adversarial imitation from observation, considering state-transitions occupancy measure. Yang et al. later improves this by minimizing the inverse dynamics disagreement instead
\cite{yang2019imitation}. Yet, all of the methods are flawed for the requirement of a great number of expert samples to train adversarial
networks, as well as the inherent complexity of adversarial imitation learning algorithms. Torabi et al. propose a combination of linear
quadratic regulators and adversarial methods, using quadratic terms to represent the cost to achieve a higher sample efficiency.
Hong et al. \cite{DBLP:journals/corr/abs-1806-10019} resort to adversarial fashion with a shared state transition reward, allowing a more efficient active exploration in the environment.

\section{Preliminaries}

\subsection{Notations}
Our NEARL framework consists of a state-to-state meta policy and a low-level inverse dynamics model.
Therefore our MDP is defined differently, using a 6-tuple $M=\{S, A, \hat{S}^{'}, P, r, \gamma \}$
, where $S$ denotes agent's state space, $\hat{S}^{'}$ denotes the space of predicted next state, $A$ is agent's action space,
$P:S\times A\rightarrow [0,1]$ is the transition probability function, with $P(s_{t+1}|s_t,a_t)$ being the
probability of transitioning from state $s_t$ to $s_{t+1}$ after taking action
$a_t$. $r: S \times S \rightarrow \mathbb{R}$ is the immediate reward and $\gamma$ is a discount factor.

\begin{algorithm}[thb]
  \caption{Non-Explicit Action Reinforcement Learning Framework}
  \begin{algorithmic}\label{alg:main}
    \REQUIRE policy $\pi_{\theta_{\pi}}$, inverse dynamics model
    $\mathcal{I}_{\theta_{\mathcal{I}}}$, environment $\mathcal{E}$.
    \FOR {$i=0,1,2,...$}
    \STATE Execute combined policy $\mathcal{I}\odot\pi$ in environment $\mathcal{E}$, and collect the trajectories $\tau_i$
    \IF {condition of optimizing  $\pi$ is satisfied}
    \STATE Update the policy parameters $\theta_{\pi}$
    \ENDIF
    \IF {condition of optimizing  $\mathcal{I}$ is satisfied}
    \STATE Update the inverse dynamics model parameters $\theta_{\mathcal{I}}$
    \ENDIF
    \IF {condition of maximizing  $\mathcal{E}$ reward is satisfied}
    \STATE Update the joint parameters $\theta=\theta_{\pi} \cup \theta_{\mathcal{I}}$
    \ENDIF
    \ENDFOR
  \end{algorithmic}
\end{algorithm}
\subsection{The Original NEARL Framework}
Original Non-Explicit Action Reinforcement Learning framework is illustrated in
Fig. \ref{fig:model}(a). The NEARL is a hierarchical framework. Meta policy $\pi$ tries to choose the next optimal state, while low-level IDM tries to find the optimal action for transition. We define trajectory as $\tau = (s_0, \hat{s}_{1}, a_0,...,s_{T+1})$ and our objective is to maximize the cumulative reward $J(\pi)=\mathbb{E}_{\tau \sim \mathcal{I}\odot\pi}[R(\tau)]$, where $\mathcal{I}\odot\pi$ is the combined policy. We also try to make sure that there will be as many reachable states as possible. If $\pi_{\theta_\pi}(s)$ is defined to be deterministic and inverse dynamics model $\mathcal{I}_{\theta_I}(a_t|s_t, \hat{s_{t+1}})$ to be stochastic, the policy gradient is derived as
\begin{equation}
  \label{equ:pg}
  \begin{aligned}
    &\nabla_\theta{\mathbb{E}}_{\tau \sim \mathcal{I}\odot\pi(\tau)} \left[ {R(\tau)} \right] = \\
    &\underset{\tau \sim \mathcal{I}\odot\pi}{\mathbb{E}} \left[ \sum_{t=0}^T \nabla_\theta \log  \left\{ \mathcal{I}_{\theta_I}(a_t|s_t, \hat{s_{t+1}}) \pi_{\theta_P}(\hat{s_{t+1}}|s_t) \right\} R(\tau) \right] 
  \end{aligned}
\end{equation}

Its algorithmic details are shown in Alg.\ref{alg:main}. Firstly, the combined policy $\mathcal{I}\odot\pi$ interacts with the environments and collects trajectories. Then meta policy $\pi$ and subsequently IDM are updated with the collected $(s,s')$ pair and $(s,s',a)$ pair respectively by using supervised learning. Then the combined policy $\mathcal{I}\odot\pi$ is updated using policy gradient with Eq.\ref{equ:pg} to maximize the cumulative reward. The original NEARL iteratively executes these steps if certain task-wise conditions are met.  However, the state transitions $(s,s')$ collected during interaction is not optimal, the direct supervised learning update of meta policy only ensures the reachability but not optimality of the predicted next state, which may lead to instability.

\subsection{The $ \rm \textit{PID}^2$ Algorithm}
The proposed NEARL framework confronts us with a multiple objective optimization problem since we need to maximize the cumulative reward while ensuring the output state of policy $\pi$ is reachable
and, this leads to more challenges in the training process compared to traditional RL.
In order to stabilize the training process, a generative adversarial network is introduced, together with the information bottleneck to regularize the discriminator, to encourage the model to focus on the most significant features.
Specifically, we use variational discriminator bottleneck \cite{peng2018variational}.
Under this design, the expected state transition $x\triangleq (s,\hat{s'})$ generated by $\pi$, is output to an encoder $E(z|x)$, and a constraint $I_c$ is further applied as the mutual information upper bound between the
encoding and input. 
Using variational approximation, the objective of GAN can be represented as:
\begin{equation}
  \label{equ:vgan}
  \begin{aligned}
    J(D, E) = \ & \underset{D, E}{\text{min}}\ \underset{\beta \geq 0}{\text{max}} \;
    \mathbb{E}_{\mathbf{x} \sim p^*(\mathbf{x})} \left[\mathbb{E}_{\mathbf{z} \sim E(\mathbf{z} | \mathbf{x})} \left[-\log  \left(D(\mathbf{z})\right) \right]\right] \\
    & + \mathbb{E}_{\mathbf{x} \sim \pi(s)} \left[\mathbb{E}_{\mathbf{z} \sim E(\mathbf{z} | \mathbf{x})} \left[-\log  \left(1 - D(\mathbf{z})\right) \right]\right] \\
    & + \beta \left(\mathbb{E}_{\mathbf{x} \sim \tilde{p}(\mathbf{x})}\left[\mathrm{KL}\left[E(\mathbf{z}|\mathbf{x}) \middle|| r(\mathbf{z}) \right] \right] - I_c \right)
  \end{aligned}
\end{equation}
where the $p^*(x)$ denotes the distribution of real state transitions. 
$\tilde{p}\triangleq p^*+\pi$ denotes a mixture distribution and
$r(z)$ is a standard Gaussian prior distribution.

The proposed Policy-Inverse Dynamics-Discriminator ($\textit{PID}^2$) algorithm is shown in Alg.\ref{alg:main}.
In the beginning, the combined policy $\mathcal{I}\odot\pi$ is used to collect trajectories $\tau$ and $\mathcal{D}_{GAN}\triangleq\{(s,s')\}\subseteq\tau$ serves as true state transition samples of the discriminator. Under the GAN framework, the discriminator $\mathcal{D}$ and policy $\pi$ are updated with Eq. \ref{equ:vgan}.

The inverse dynamics model $\mathcal{I}$ should be updated cautiously, since once the $\mathcal{I}$ is changed, the environmental state transition is also changed and it easily leads to instability during training. To tackle the potential instability, we set an inverse dynamics update threshold $ [\sigma_\mathcal{I}]$ which triggers update only if the inverse dynamics model loss exceeds the threshold. 
The threshold $ [\sigma_\mathcal{I}]$ can be adjusted according to the task and required control accuracy.
The IDM is trained by dataset $\mathcal{D}_{\mathcal{I}}\triangleq\{(s,s',a)\}\subseteq\tau$ with supervised learning:
\begin{equation}
  \label{equ:ID}
  L_{\mathcal{I}} = -\sum_{(s_t, a_t, s_{t+1}) \in \tau} {\log\mathcal{I}_{\theta_{I}}(a_t|s_t,s_{t+1})}
\end{equation}

\begin{algorithm}[t!]
  \caption{The $\rm \textit{PID}^2$ Algorithm}
  \label{alg:main}
  \begin{algorithmic}[1]
    \REQUIRE policy $\pi_{\theta_{\pi}}$, inverse dynamics model
    $\mathcal{I}_{\theta_{\mathcal{I}}}$, environment $\mathcal{E}$. discriminator
    $\mathcal{D_\phi}$, inverse dynamics loss threshold $[\sigma_\mathcal{I}]$,
    optional state-only demonstration $\mathcal{D}_{demo}$, optional prior
    constraints $\mathcal{C}$
    \IF {$\mathcal{D}_{demo}$ available}
    \STATE Pre-train the policy parameters $\theta_{\pi}$ with Equation 1
    \ELSE
    \STATE Randomly initialize policy parameters $\theta_{\pi}$
    \ENDIF
    \FOR {$i=0,1,2,...$}
    \STATE Execute  combined policy $\mathcal{I}\odot\pi$ in environment
    $\mathcal{E}$ with constraints $\mathcal{C}$, and collect the trajectories $\tau_i$
    \STATE Update $\phi$ and $\theta_{\pi}$ using Equation 2
    \IF {Inverse dynamics model loss $\mathcal{I}_{\theta_{\mathcal{I}}} \geq [\sigma_\mathcal{I}]$}
    \STATE Update $\theta_{\mathcal{I}}$ using Equation 4
    \ENDIF
    \STATE Update $\theta_{\pi} \cup \theta_{\mathcal{I}}$ using Equation 5
    \ENDFOR
  \end{algorithmic}
\end{algorithm}

The algorithm is also required to maximize the environmental rewards like traditional RL, which could be achieved via the aforementioned policy gradient (Eq. 
\ref{equ:pg}) to optimize the combined policy.

Besides, imitation learning with state-only demonstrations could be easily integrated with our algorithm. Given state-only expert data $\mathcal{D}_{demo}=\{(s_i,s_{i+1})\}$, we could apply pre-training process using behavior cloning (Eq.\ref{equ:BC}).
Except for behavior cloning, the demonstrations could be appended into $\mathcal{D}_{GAN}$ in order to encourage the policy $\pi$ to mimic the expert.
\begin{equation}
\label{equ:BC}
L_{BC} = \sum_{(s_t, s_{t+1}) \in \mathcal{D}_{demo}} {|s_{t+1} - \pi(s_t)|}^2
\end{equation}

Finally, the state-related prior domain knowledge and constraints can be directly applied to the policy. For example, states can be divided into controllable state $S{c}$ (e.g. joint angles) and observable but non-controllable state $S{o}$ (e.g. visual data). By using this prior knowledge, merely controllable state $S{c}$ can be used to train IDM.

\section{experimental results}
We conduct several experiments to estimate our algorithm 
and its combination with imitation learning.
We first verify the validity of the NEARL framework and $\textit{PID}^2$ in simulation environments. 
After that, we conduct experiments to elucidate that our algorithms can be effectively combined with imitation learning, merely using state-only demonstrations.

\subsection{$\textit{PID}^2$ Evaluation}
We evaluate our algorithm on several robotic control tasks via OpenAI gym environments \cite{gym} simulated by MuJoCo physical engine \cite{todorov2012mujoco}.
The simulation
tasks include Ant, HalfCheetah, Walker2d and Hopper.
We evaluate traditional RL, original NEARL, $\textit{PID}^2$ (w/o information bottleneck) and $\textit{PID}^2$ (with information bottleneck).  The same neural networks are used for all algorithms. For more details, policy network, inverse dynamics model, and discriminator are set to fully-connected neural networks with hidden layer size of $32 \times 32$ and IDM outputs a Gaussian distribution. During the training process, PPO
\cite{schulman2017proximal} is used to train the combined policy with 8 million steps.

\begin{figure*}[!tbh]
  \centering
  \includegraphics[width=17.5cm]{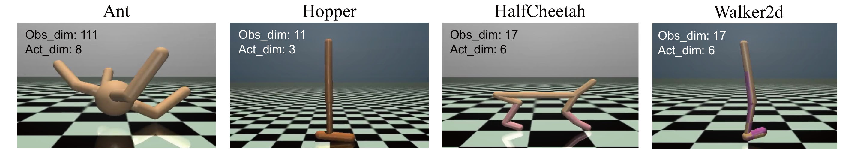}
  \caption{Demonstration of four MuJoCO environments. Dimension of observation
    space and action space are displayed respectively.}
  \label{fig:environment}
  \centering
  \includegraphics[width=17.5cm]{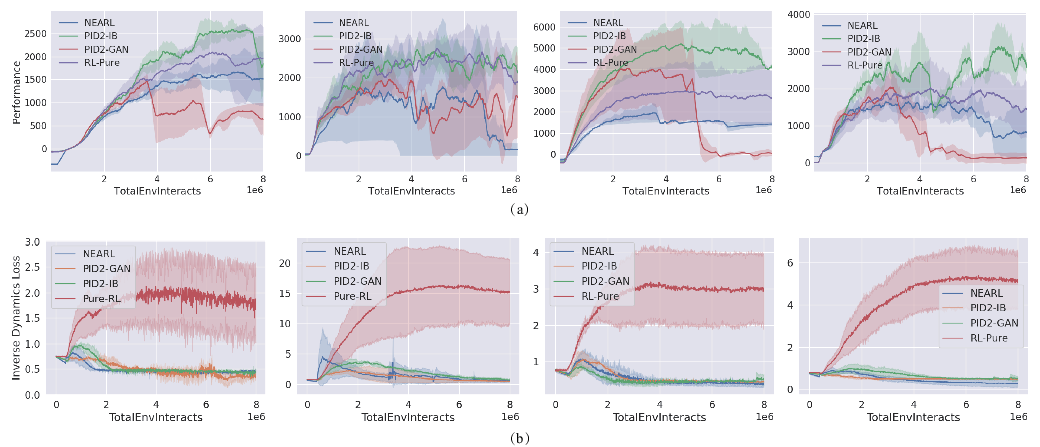}
  \caption{(a) Different algorithms' performance with four simulation tasks. NEARL represents the original NEARL framework. PID2-IB represents the $\textit{PID}^2$ with information bottleneck while the PID2-GAN represents the $\textit{PID}^2$ with only GAN structure but no information bottleneck. The RL-Pure is the PPO algorithm. (b) Inverse dynamics
    loss of four algorithms.}
  \label{fig:loss_reward}
\end{figure*}

The experimental results for each task are illustrated in Fig. \ref{fig:environment} and Fig. \ref{fig:loss_reward}.
Fig. \ref{fig:loss_reward}(a) plots the learning curves for all methods in simulation environments and Fig. \ref{fig:loss_reward}(b) displays their inverse dynamics loss (mean square error).
Due to the existence of multiple optimized objectives, the performance of our original NEARL framework is lower than traditional RL. However, the algorithm performance improves to a similar level with traditional RL after applying GAN with information bottleneck, and in some tasks even better. It may be attributed to the estimation of reachable states, which benefits training the deep neural network.

As for simple tasks, the performance of $\textit{PID}^2$ with information bottleneck is quite the same as that of $\textit{PID}^2$ without information bottleneck. But when the state representation of the environment has a lot of redundant information (e.g., Ant environment with state dimension of 111), the $\textit{PID}^2$ with information bottleneck significantly outperforms the one without it.
This demonstrates that the variational information bottleneck has shown its benefits in extracting the most effective features among redundant states.
In the original NEARL method, adversarial learning is not used. Instead, we directly apply supervised constraints to the output of the policy network by using on-policy state-transition samples. This could lead to instability and bad performance during the training process since it will disturb the policy gradient update in the following steps. 

The accuracy of inverse dynamics model is shown in Fig.\ref{fig:loss_reward}(b).
Serving as a baseline, traditional RL does not put any constraint on inverse
dynamics loss, while in our NEARL-based approaches, we set a threshold
$[\sigma_\mathcal{I}] = 0.5$. As demonstrated, inverse dynamics loss is effectively
limited and much smaller than the baseline's, which guarantees required control
precision.

\subsection{$\textit{PID}^2$ with Imitation Learning}
\subsubsection{Performance Comparison}
Because NEARL-based policy outputs an optimal reachable state rather than an explicit action, our $\textit{PID}^2$ algorithm can effectively combine state-only expert demonstrations. Here, the $\textit{PID}^2$ algorithm together with two other algorithms (state-only imitation learning (SOIL) \cite{radosavovic2020state} and demo augmented policy gradient (DAPG)\cite{rajeswaran2017learning}) are evaluated in the aforementioned environments. DAPG is firstly pre-trained with behavior cloning using state-action pair demonstration, then fine-tuned using augmented loss of RL policy gradient and behavior cloning gradient. In the SOIL algorithm, an IDM is trained first, then the state-only demonstrations are completed with inferred actions, afterwards optimization method is adopted for training similar to DAPG.
We found
that in the DAPG algorithm, pre-training plays a crucial role in increasing
performance, while the SOIL cannot be pre-trained directly due to the absence of
action in the early phase, which leads to slow convergence of the algorithm. To
be fair, we adopted Initial State Distribution \cite{peng2018deepmimic} for SOIL
and $PID^2$ algorithm. In the experiment, pre-training is performed for 20
episodes, then normal policy search methods. Besides, prior knowledge of
controllable state is also exploited for IDM.
\begin{figure}[!hb]
  \flushleft
  \includegraphics[width=8.7cm]{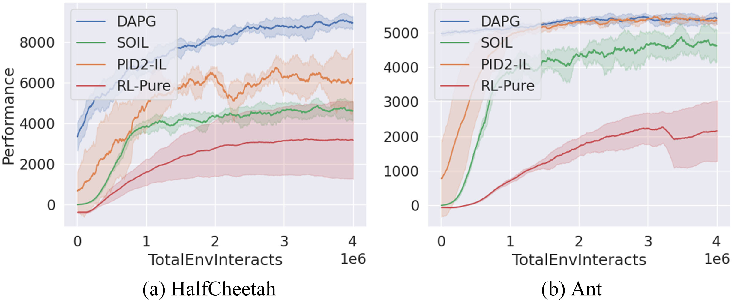}
  \caption{Performance comparison of four different algorithms combining imitation learning, with Pure RL serving as baseline.}
  \label{fig:exp2}
\end{figure}

Fig.\ref{fig:exp2} shows the performance of four approaches in two simulation
environments. Compared to RL algorithm without imitation learning, the
performance of three other algorithms are enhanced remarkably. Since
ground-truth action is available for behavior cloning, DAPG excels the other
three algorithms. As for SOIL, in its early phase, data is not sufficient to
train the IDM and the poor accuracy of its IDM heavily affects its performance.

\mathchardef\mhyphen="2D
Besides, in our $\textit{PID}^2\mhyphen IL$ algorithm, meta policy is pretrained with state-only demonstrations and combined policy is optimized by multiple means. Hence our algorithm outperforms SOIL.

\subsubsection{Ablation Study}
In the aforementioned experiments, we used prior human knowledge and
hand-engineered features (controllable states in observation space) for IDM's
input. But occasionally it is hard to tell which part of the state
representation could have more significant influence on inverse dynamics. Here
we investigate the performance of our algorithms with wrong or no human prior
knowledge.

\begin{figure}[!thb]
  \flushleft
  \includegraphics[width=8.7cm]{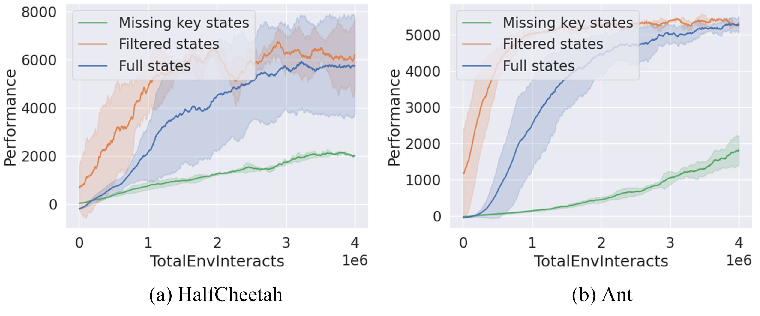}
  \caption{ Influence of prior knowledge to the performance in 3 cases:
    correct prior knowledge (correctly filtered  states), wrong prior knowledge (missing key states) and no prior knowledge (unfiltered full states).}
  \label{fig:exp3}
\end{figure}

The results are demonstrated in Fig.\ref{fig:exp3}. We compared the performance of our method with full states (unfiltered states) and wrongly filtered states (missing half of key states). The results demonstrate that the performance of full state $PID^2 \mhyphen IL$ has a noticeable decline in both environments in terms of convergence rate and a slight decline in its cumulative rewards.
This is due to lack of data to train a more complicated full state IDM which finally may lead to trajectory distribution shift. Besides, if the state representation is wrongly filtered like missing key state, the performance may dramatically decrease, since the IDM cannot learn an effective feature mapping, leaving the policy gradient the only part of the algorithm to optimize the policy. In such a case, behavior cloning cannot contribute to policy optimization but instead disturb the training process.
Comparing the full states $PID^2 \mhyphen IL$ and missing key states $PID^2 \mhyphen IL$, it is revealed that if we do not have perfect prior knowledge, it would be better to keep the redundancy in state representation rather than incorrectly filtering the states since the later may even lead to a even worse performance. Maybe an automated learning process of feature mapping exists as a better solution.

\section{Conclusion}

We proposed a new Non-Explicit Action Reinforcement Learning (NEARL) framework and the $\textit{PID}^2$ algorithm.
Our method adopts the hierarchical structure, iteratively training a state-only policy and IDM.
Also, adversarial learning and the information bottleneck are introduced to stabilize the training process. Our method shows equivalent performance in several robotic simulation tasks with traditional RL and is able to constrain the inverse dynamics loss according to the task to ensure the state reachability. It can also be integrated with state-only expert demonstrations, showing state of the art performance.

Additionally, our state-to-state policy inherently integrates an environmental
model. The model enables our method not only to be used in one step MDP, but
also sequence learning. 
In our future work, we will improve the method by combining the advanced sequence learning methods to learn action primitives \cite{akbulut2020adaptive}.
Yet, we here still need to point out that our algorithm shows instability in high-dimensional environments, hence it should  be carefully applied to complicated tasks or the physical world. We will explore other methods like self-attention\cite{wang2017residual} to improve its robustness and expand its application in sim-to-real transfer.

%%%%%%%%%%%%%%%%%%%%%%%%%%%%%%%%%%%%%%%%%%%%%%%%%%%%%%%%%%%%%%%%%%%%%%%%%%%%%%%%
\bibliographystyle{IEEEtran}
\bibliography{IEEEabrv,root}

% Generated by IEEEtran.bst, version: 1.14 (2015/08/26)
\begin{thebibliography}{10}
\providecommand{\url}[1]{#1}
\csname url@samestyle\endcsname
\providecommand{\newblock}{\relax}
\providecommand{\bibinfo}[2]{#2}
\providecommand{\BIBentrySTDinterwordspacing}{\spaceskip=0pt\relax}
\providecommand{\BIBentryALTinterwordstretchfactor}{4}
\providecommand{\BIBentryALTinterwordspacing}{\spaceskip=\fontdimen2\font plus
\BIBentryALTinterwordstretchfactor\fontdimen3\font minus
  \fontdimen4\font\relax}
\providecommand{\BIBforeignlanguage}[2]{{%
\expandafter\ifx\csname l@#1\endcsname\relax
\typeout{** WARNING: IEEEtran.bst: No hyphenation pattern has been}%
\typeout{** loaded for the language `#1'. Using the pattern for}%
\typeout{** the default language instead.}%
\else
\language=\csname l@#1\endcsname
\fi
#2}}
\providecommand{\BIBdecl}{\relax}
\BIBdecl

\bibitem{silver2017mastering}
D.~Silver, T.~Hubert, J.~Schrittwieser, I.~Antonoglou, M.~Lai, A.~Guez,
  M.~Lanctot, L.~Sifre, D.~Kumaran, T.~Graepel \emph{et~al.}, ``Mastering chess
  and shogi by self-play with a general reinforcement learning algorithm,''
  \emph{arXiv preprint arXiv:1712.01815}, 2017.

\bibitem{akkaya2019solving}
I.~Akkaya, M.~Andrychowicz, M.~Chociej, M.~Litwin, B.~McGrew, A.~Petron,
  A.~Paino, M.~Plappert, G.~Powell, R.~Ribas \emph{et~al.}, ``Solving rubik's
  cube with a robot hand,'' \emph{arXiv preprint arXiv:1910.07113}, 2019.

\bibitem{achiam2017constrained}
J.~Achiam, D.~Held, A.~Tamar, and P.~Abbeel, ``Constrained policy
  optimization,'' \emph{arXiv preprint arXiv:1705.10528}, 2017.

\bibitem{miryoosefi2019reinforcement}
S.~Miryoosefi, K.~Brantley, H.~Daume~III, M.~Dudik, and R.~E. Schapire,
  ``Reinforcement learning with convex constraints,'' in \emph{Advances in
  Neural Information Processing Systems}, 2019, pp. 14\,093--14\,102.

\bibitem{argall2009survey}
B.~D. Argall, S.~Chernova, M.~Veloso, and B.~Browning, ``A survey of robot
  learning from demonstration,'' \emph{Robotics and autonomous systems},
  vol.~57, no.~5, pp. 469--483, 2009.

\bibitem{hussein2017imitation}
A.~Hussein, M.~M. Gaber, E.~Elyan, and C.~Jayne, ``Imitation learning: A survey
  of learning methods,'' \emph{ACM Computing Surveys (CSUR)}, vol.~50, no.~2,
  pp. 1--35, 2017.

\bibitem{peng2018deepmimic}
X.~B. Peng, P.~Abbeel, S.~Levine, and M.~van~de Panne, ``Deepmimic:
  Example-guided deep reinforcement learning of physics-based character
  skills,'' \emph{ACM Transactions on Graphics (TOG)}, vol.~37, no.~4, pp.
  1--14, 2018.

\bibitem{peng2018sfv}
X.~B. Peng, A.~Kanazawa, J.~Malik, P.~Abbeel, and S.~Levine, ``Sfv:
  Reinforcement learning of physical skills from videos,'' \emph{ACM
  Transactions on Graphics (TOG)}, vol.~37, no.~6, pp. 1--14, 2018.

\bibitem{levine2013guided}
S.~Levine and V.~Koltun, ``Guided policy search,'' in \emph{International
  Conference on Machine Learning}, 2013, pp. 1--9.

\bibitem{levine2014learning}
S.~Levine and P.~Abbeel, ``Learning neural network policies with guided policy
  search under unknown dynamics,'' in \emph{Advances in Neural Information
  Processing Systems}, 2014, pp. 1071--1079.

\bibitem{hester2013texplore}
T.~Hester and P.~Stone, ``Texplore: real-time sample-efficient reinforcement
  learning for robots,'' \emph{Machine learning}, vol.~90, no.~3, pp. 385--429,
  2013.

\bibitem{jong2007model}
N.~K. Jong and P.~Stone, ``Model-based function approximation in reinforcement
  learning,'' in \emph{Proceedings of the 6th international joint conference on
  Autonomous agents and multiagent systems}, 2007, pp. 1--8.

\bibitem{deisenroth2011pilco}
M.~Deisenroth and C.~Rasmussen, ``Pilco: A model-based and data-efficient
  approach to policy search.'' 01 2011, pp. 465--472.

\bibitem{wahlstrom2015pixels}
N.~Wahlström, T.~B. Schön, and M.~P. Deisenroth, ``From pixels to torques:
  Policy learning with deep dynamical models,'' 2015.

\bibitem{depeweg2017learning}
S.~Depeweg, J.~M. Hernández-Lobato, F.~Doshi-Velez, and S.~Udluft, ``Learning
  and policy search in stochastic dynamical systems with bayesian neural
  networks,'' 2017.

\bibitem{moerland2017learning}
T.~M. Moerland, J.~Broekens, and C.~M. Jonker, ``Learning multimodal transition
  dynamics for model-based reinforcement learning,'' 2017.

\bibitem{sutton1991dyna}
R.~S. Sutton, ``Dyna, an integrated architecture for learning, planning, and
  reacting,'' \emph{ACM Sigart Bulletin}, vol.~2, no.~4, pp. 160--163, 1991.

\bibitem{watter2015embed}
M.~Watter, J.~Springenberg, J.~Boedecker, and M.~Riedmiller, ``Embed to
  control: A locally linear latent dynamics model for control from raw
  images,'' in \emph{Advances in neural information processing systems}, 2015,
  pp. 2746--2754.

\bibitem{chua2018deep}
K.~Chua, R.~Calandra, R.~McAllister, and S.~Levine, ``Deep reinforcement
  learning in a handful of trials using probabilistic dynamics models,'' 2018.

\bibitem{williams2017information}
G.~{Williams}, N.~{Wagener}, B.~{Goldfain}, P.~{Drews}, J.~M. {Rehg},
  B.~{Boots}, and E.~A. {Theodorou}, ``Information theoretic mpc for
  model-based reinforcement learning,'' in \emph{2017 IEEE International
  Conference on Robotics and Automation (ICRA)}, 2017, pp. 1714--1721.

\bibitem{sutton2018reinforcement}
R.~S. Sutton and A.~G. Barto, \emph{Reinforcement learning: An
  introduction}.\hskip 1em plus 0.5em minus 0.4em\relax MIT press, 2018.

\bibitem{giusti2015machine}
A.~Giusti, J.~Guzzi, D.~C. Cire{\c{s}}an, F.-L. He, J.~P. Rodr{\'\i}guez,
  F.~Fontana, M.~Faessler, C.~Forster, J.~Schmidhuber, G.~Di~Caro
  \emph{et~al.}, ``A machine learning approach to visual perception of forest
  trails for mobile robots,'' \emph{IEEE Robotics and Automation Letters},
  vol.~1, no.~2, pp. 661--667, 2015.

\bibitem{bojarski2016end}
M.~Bojarski, D.~Del~Testa, D.~Dworakowski, B.~Firner, B.~Flepp, P.~Goyal, L.~D.
  Jackel, M.~Monfort, U.~Muller, J.~Zhang \emph{et~al.}, ``End to end learning
  for self-driving cars,'' \emph{arXiv preprint arXiv:1604.07316}, 2016.

\bibitem{bain1995framework}
M.~Bain and C.~Sammut, ``A framework for behavioural cloning.'' in
  \emph{Machine Intelligence 15}, 1995, pp. 103--129.

\bibitem{codevilla2019exploring}
F.~Codevilla, E.~Santana, A.~M. L{\'o}pez, and A.~Gaidon, ``Exploring the
  limitations of behavior cloning for autonomous driving,'' in
  \emph{Proceedings of the IEEE International Conference on Computer Vision},
  2019, pp. 9329--9338.

\bibitem{ho2016generative}
J.~Ho and S.~Ermon, ``Generative adversarial imitation learning,'' in
  \emph{Advances in neural information processing systems}, 2016, pp.
  4565--4573.

\bibitem{peng2018variational}
X.~B. Peng, A.~Kanazawa, S.~Toyer, P.~Abbeel, and S.~Levine, ``Variational
  discriminator bottleneck: Improving imitation learning, inverse rl, and gans
  by constraining information flow,'' \emph{arXiv preprint arXiv:1810.00821},
  2018.

\bibitem{torabi2018generative}
F.~Torabi, G.~Warnell, and P.~Stone, ``Generative adversarial imitation from
  observation,'' \emph{arXiv preprint arXiv:1807.06158}, 2018.

\bibitem{torabi2019recent}
------, ``Recent advances in imitation learning from observation,'' 2019.

\bibitem{sermanet2018time}
P.~Sermanet, C.~Lynch, Y.~Chebotar, J.~Hsu, E.~Jang, S.~Schaal, S.~Levine, and
  G.~Brain, ``Time-contrastive networks: Self-supervised learning from video,''
  in \emph{2018 IEEE International Conference on Robotics and Automation
  (ICRA)}.\hskip 1em plus 0.5em minus 0.4em\relax IEEE, 2018, pp. 1134--1141.

\bibitem{aytar2018playing}
Y.~Aytar, T.~Pfaff, D.~Budden, T.~Paine, Z.~Wang, and N.~de~Freitas, ``Playing
  hard exploration games by watching youtube,'' in \emph{Advances in Neural
  Information Processing Systems}, 2018, pp. 2930--2941.

\bibitem{radosavovic2020state}
I.~Radosavovic, X.~Wang, L.~Pinto, and J.~Malik, ``State-only imitation
  learning for dexterous manipulation,'' \emph{arXiv preprint
  arXiv:2004.04650}, 2020.

\bibitem{torabi2018behavioral}
F.~Torabi, G.~Warnell, and P.~Stone, ``Behavioral cloning from observation,''
  in \emph{Proceedings of the 27th International Joint Conference on Artificial
  Intelligence}.\hskip 1em plus 0.5em minus 0.4em\relax AAAI Press, 2018, pp.
  4950--4957.

\bibitem{merel2017learning}
J.~Merel, Y.~Tassa, D.~TB, S.~Srinivasan, J.~Lemmon, Z.~Wang, G.~Wayne, and
  N.~Heess, ``Learning human behaviors from motion capture by adversarial
  imitation,'' \emph{arXiv preprint arXiv:1707.02201}, 2017.

\bibitem{yang2019imitation}
C.~Yang, X.~Ma, W.~Huang, F.~Sun, H.~Liu, J.~Huang, and C.~Gan, ``Imitation
  learning from observations by minimizing inverse dynamics disagreement,'' in
  \emph{Advances in Neural Information Processing Systems}, 2019, pp. 239--249.

\bibitem{DBLP:journals/corr/abs-1806-10019}
\BIBentryALTinterwordspacing
Z.~Hong, T.~Fu, T.~Shann, Y.~Chang, and C.~Lee, ``Adversarial exploration
  strategy for self-supervised imitation learning,'' \emph{CoRR}, vol.
  abs/1806.10019, 2018. [Online]. Available:
  \url{http://arxiv.org/abs/1806.10019}
\BIBentrySTDinterwordspacing

\bibitem{gym}
G.~Brockman, V.~Cheung, L.~Pettersson, J.~Schneider, J.~Schulman, J.~Tang, and
  W.~Zaremba, ``Openai gym,'' 2016.

\bibitem{todorov2012mujoco}
E.~Todorov, T.~Erez, and Y.~Tassa, ``Mujoco: A physics engine for model-based
  control,'' in \emph{2012 IEEE/RSJ International Conference on Intelligent
  Robots and Systems}.\hskip 1em plus 0.5em minus 0.4em\relax IEEE, 2012, pp.
  5026--5033.

\bibitem{schulman2017proximal}
J.~Schulman, F.~Wolski, P.~Dhariwal, A.~Radford, and O.~Klimov, ``Proximal
  policy optimization algorithms,'' \emph{arXiv preprint arXiv:1707.06347},
  2017.

\bibitem{rajeswaran2017learning}
A.~Rajeswaran, V.~Kumar, A.~Gupta, G.~Vezzani, J.~Schulman, E.~Todorov, and
  S.~Levine, ``Learning complex dexterous manipulation with deep reinforcement
  learning and demonstrations,'' \emph{arXiv preprint arXiv:1709.10087}, 2017.

\bibitem{akbulut2020adaptive}
M.~T. Akbulut, M.~Y. Seker, A.~E. Tekden, Y.~Nagai, E.~Oztop, and E.~Ugur,
  ``Adaptive conditional neural movement primitives via representation sharing
  between supervised and reinforcement learning,'' \emph{arXiv preprint
  arXiv:2003.11334}, 2020.

\bibitem{wang2017residual}
F.~Wang, M.~Jiang, C.~Qian, S.~Yang, C.~Li, H.~Zhang, X.~Wang, and X.~Tang,
  ``Residual attention network for image classification,'' in \emph{Proceedings
  of the IEEE conference on computer vision and pattern recognition}, 2017, pp.
  3156--3164.

\end{thebibliography}

%\end{CJK*}
\end{document}